\pdfoutput=1

\documentclass[11pt]{article}

\usepackage{acl}

\usepackage{times}
\usepackage{latexsym}
\usepackage{comment}
\usepackage{subcaption}
\usepackage{graphicx}
\usepackage{footnote}
\usepackage{multirow}
\usepackage{tabulary}
\usepackage{multicol}
\usepackage[detect-none]{siunitx}
\usepackage[T1]{fontenc}
\usepackage{xspace}
\usepackage{color, colortbl}

\usepackage{microtype}
\usepackage[normalem]{ulem}
\usepackage{booktabs}
\usepackage[colorinlistoftodos]{todonotes}
\usepackage{enumitem}
\usepackage{xspace}
\usepackage{amssymb}
\usepackage{pifont}
\usepackage{algorithm,algpseudocode}
\usepackage{amsmath}
\usepackage{amsfonts}
\usepackage{amssymb}
\usepackage{listings}
\usepackage{tikz}

\newcolumntype{L}{>{\centering\arraybackslash}m{4.5cm}}
\usepackage{makecell}

\usepackage[utf8]{inputenc}

\usepackage{microtype}

\newcommand{\nilm}{{\fontfamily{lmtt}\selectfont NILM}\xspace}

\title{What do Large Language Models Learn beyond Language?}

\author{Avinash Madasu \unskip\enspace{\rm}\enspace Shashank Srivastava\\
  UNC Chapel Hill\\ 
  \texttt{\{avinashm,ssrivastava\}@cs.unc.edu}
}

\begin{document}
\maketitle
\begin{abstract}
 Large language models (LMs) have rapidly become a mainstay in Natural Language Processing. These models are known to acquire rich linguistic knowledge from training on large amounts of text. In this paper, we investigate if pre-training on text also confers these models with helpful `inductive biases' for non-linguistic reasoning. On a set of 19 diverse non-linguistic tasks involving quantitative computations, recognizing regular expressions and reasoning over strings. We find that pretrained models significantly outperform comparable non-pretrained neural models. This remains true also in experiments with training non-pretrained models with fewer parameters to account for model regularization effects. We further explore the effect of text domain on LMs by pretraining models from text from different domains and provenances. Our experiments surprisingly reveal that the positive effects of pre-training persist even when pretraining on multi-lingual text or computer code, and even for text generated from synthetic languages. Our findings suggest a hitherto unexplored deep connection between pre-training and inductive learning abilities of language models\footnote{\url{https://github.com/avinashsai/NILM}}.
\end{abstract}

\section{Introduction}
Pretrained Language Models (LMs) 
have shown singular succcess on a range of natural language understandings tasks, to the extent that they have become foundational for contemporary NLP systems. 
Several works have investigated why pretraining works so well \cite{warstadt-etal-2019-investigating, zhao-etal-2020-quantifying}. In particular, studies have shown that the pretrained LMs like BERT capture linguistic knowledge about syntax \cite{lin2019open, wu-etal-2020-perturbed}, semantics \cite{vulic-etal-2020-probing, vulic-etal-2020-multi} and morphology \cite{hofmann2020dagobert, hofmann-etal-2021-superbizarre}. In fact, \newcite{tenney-etal-2019-bert} demonstrated that learned representations in pretrained LMs even internally reflect the classical NLP pipeline. Since most NLP benchmarks such as SuperGLUE \cite{wang2019superglue} naturally are focused on tasks such as textual entailment and reading comprehension that require linguistic knowledge and reasoning, it is unsurprising that LMs have achieved strong results on these tasks. On the other hand, little work so far has explored the abilities of pretrained LMs for learning non-linguistic tasks. 

\begin{figure}[t]
\includegraphics[width=\columnwidth]{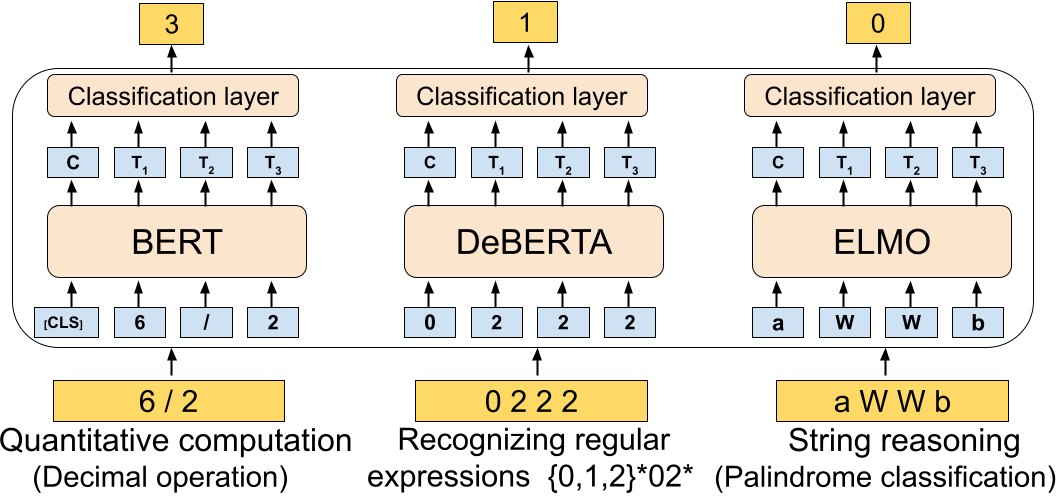}
\caption{We investigate the effect of pretraining of languages models on learning non-linguistic 
tasks using three task paradigms involving symbolic reasoning.} 
\label{fig:taskdef}
\end{figure}


\begin{table*}[]
\footnotesize
\centering
\begin{tabular}{|c|L|c|c|c|} \hline
 \thead{Task}  & \thead{Input Eg.} & \thead{Output   Eg.} & \thead{Classes} & \thead{Input range} \\
 \hline
 Odd classification & \textit{4210} & 0 & 0 - 1 & [1, 20000]  \\
 \hline
 Even classification & \textit{4210} & 1 & 0 - 1 & [1, 20000]  \\
 \hline
  Odd even classification & \textit{4210 even} & 1 & 0 - 1 & [1, 20000]  \\
 \hline
Decimal operation  & \textit{872 / 436} & 2 & 0 - 9 &  [1, 10000] \\
\hline
Decimal \& word operation & \textit{four / 2} & 2 & 0 - 9 &  [1, 10000] \\
\hline
Mean  & \textit{15,-8,15,-5,-14,-3 ?} & 0 & 0 - 9 & [-15, 15] 
\\
\hline
Median  & \textit{3,6,5,15,2,3,-6,-2,9,-3,-9,-5,-14 ?} & 2 & 0 - 9 & [-15, 15] \\
\hline
Mode  & \textit{5,9,7,0,2,5,3,3,3,0 ?} & 3 & 0 - 9 & [0, 9]  \\
\hline
Recognize \{0, 1, 2\}*02* & \textit{0 1 2 0 2 1 0 2 2 2 2} & 1 & 0 - 1 & [0, 2]  \\
\hline
Recognize AA*BB*CC*DD*EE*  & \textit{a a a a a a a b b b b c c c c c d d d e} & 1 & 0 - 1 & [a, e] \\
\hline
Palindrome classification & \textit{a W X X W a} &1 & 0 - 1 & [a-z], [A-Z] \\
\hline
Anagram classification & \textit{r G r P J h k - k h G r P J r} & 1 & 0 - 1 & [a-z],[A-Z] \\
\hline
Isogram classification & \textit{v F J o S j} & 1 & 0 - 1 & [a-z], [A-Z] \\
\hline
Tautonym classification & \textit{s t P v g - t P v g a} & 1 & 0 - 1 & [a-z], [A-Z] \\
\hline
Length of a string& \textit{t e e o} & 4 & 0 - 9 & [a-z]\\
\hline
Count of unique characters   & \textit{d e i i e e d i i d } & 3 & 0 - 9 & [a-j] \\
\hline
Parity check  & \textit{0 1 1 1 0 1 0 0 1 1 1 0} & 0 &  0 - 1 & [0, 1]\\
\hline
Vowels classification & \textit{i i v x c m o o u o} & 0 & 0 - 9 & [a-z]\\
\hline
Maximum frequent character  & \textit{j j j c j j} & 9 (j) & 0 - 9 & [a-j] \\
\hline
\end{tabular}
\caption{Description of the non-linguistic tasks with input and output examples. Classes are the class labels for each task. Input range denotes the range of the input operands in each task.}
\label{datasettable}
\end{table*}

In this paper, we explore whether pretraining on text is inherently about learning language, or if pretraining also imbues LMs with skills for symbolic manipulation and non-linguistic reasoning (for example, performing quantitative computation such as finding the median of a set of numbers, recognizing regular expressions, or identifying whether a string 
is a palindrome, as shown in Figure~\ref{fig:taskdef}). In other words, we investigate whether and how pretraining develops helpful inductive biases for non-linguistic reasoning. For this analysis, we create a set of 19 tasks from three categories of task paradigms: quantitative computation (\S  \ref{arth}), recognizing regular expressions (\S  \ref{regex}), and string reasoning (\S  \ref{string}). Figure \ref{fig:taskdef} shows an example for each category, and the full list of tasks is described in the table \ref{datasettable}. We experiment with transformer and RNN based LMs (\S  \ref{models}) for learning these tasks, and perform a comparative analysis with (non-pretrained) neural model variants from the perspective of learning metrics such as accuracy and sample efficiency.

Our experiments (\S \ref{results}) reveal that pretrained models overall perform substantially better and are more sample efficient on most tasks. However, there are significant differences and patterns in performance between task types, as well as variance between different LM architectures. Since non-pretrained models do not have the benefit of regularization that comes from pretraining, a plausible reason for the discrepancy between them and pretrained LMs might be underfitting of the non-pretrained models when trained on comparatively small dataset sizes. To account for this, we also comprehensively explore the effect of model size (\S \ref{param_count}) of non-pretrained models for both transformer and RNN architectures. We find that the discrepancy in performance remains even for smaller neural models, indicating that the differences are not simply due to a mismatch in model and data sizes.

Finally, we investigate the role that pretraining data plays in influencing task performance on non-linguistic tasks (\S  \ref{pretrainedablation}). We experiment with pretraining on different domains of text, pretraining on perturbed representations of natural language text (such as shuffled word order), pretraining on text of computer programs (no linguistic properties of natural languages), pretraining on multi-lingual and non-English text, and pretraining with synthetic text (data sampled from synthetic distributions). Our analysis reveals that the advantages of pretraining surprisingly persist with various degrees across these variations, suggesting hithertho unexplored connections between pretraining and the learning abilities of language models. Our contributions are:

\begin{itemize}[noitemsep, topsep=0pt, leftmargin=*]
    \item We compare a range of pretrained LMs and non-pretrained models on a carefully designed suite of 19 classifications tasks that require non-linguistic reasoning. 
    \item We comprehensively explore the role of the pretraining data by experimenting with models pretrained from texts with different provenances. 
    \item We establish that the positive effects of pretraining are not simply due to better model regularization by experimenting with neural models with different complexities and architectures.
\end{itemize}

\section{Related Work}

A body of work has investigated contextual word embeddings 
to determine whether they capture aspects of mathematical meaning for numbers \cite{naik-etal-2019-exploring}. 
\citet{wallace-etal-2019-nlp} probed numerical supremacy on token embeddings of contextual language models such as ELMO and BERT. \cite{thawani-etal-2021-representing} surveyed numerical understanding in NLP models using 7 sub-tasks such as measurement estimation and word problems. Our work diverges from these in exploring a richer set of tasks including harder tasks such as set operations. Further, previous methods explore mathematical reasoning tasks posed as language problems, which conflates the problems of language and mathematical learning and also makes the datasets susceptible to biases due to data collection. Our analysis circumvents both these issues by design.

Some previous works have explored the ability of RNN and Transformer architectures for learning regular languages \cite{weiss-etal-2018-practical, sennhauser-berwick-2018-evaluating, suzgun-etal-2019-evaluating, bhattamishra-etal-2020-ability}, closing brackets \cite{skachkova-etal-2018-closing}, and dynamic counting \cite{suzgun-etal-2019-lstm}. 
However, they focus on the learnability of these tasks with specific architectures, and do not look at pretrained LMs, which are our focus here.

Finally, in our discussion, we conceptually stretch the notion of inductive bias. 
The idea of inductive bias is usually associated with specific model types \cite{mccoy-etal-2020-syntax, kharitonov2021what}, architectures \cite{xu2021positional, brutzkus2021on} and regularization approaches \cite{helmbold2015inductive}. We believe that extending this to refer to learning tasks with pretrained LMs is both reasonable and useful. 

\section{NILM} \label{probe}
In this section, we describe the tasks used for our analysis, which we refer to as \nilm (measuring Non-linguistic Inductive bias in Language Models). The tasks correspond to three task paradigms: (1) quantitative computation, (2) regular expressions, and (3) string reasoning. Each task in \nilm is posed as a classification task. The descriptions for all the tasks with input and output examples, class labels and the input range are shown in Table~\ref{datasettable}. Each task has a synthetically generated dataset with train/dev/test splits\footnote{The training set size for all tasks is 10K, dev set size is 1K and test set size is 1K, except for tasks on recognizing regular expressions, where the test set size is 2K following previous work \cite{bhattamishra-etal-2020-ability}.}. To avoid biases in the datasets, relevant numbers and  strings in individual examples are uniformly sampled from the appropriate ranges. 

\subsection{Quantitative computation} \label{arth}
This task paradigm focuses on tasks involving arithmetic and set statistics. \\
\textbf{Odd classification.} 
Classify if a number is odd. \\
\textbf{Even classification.} 
Classify if a number is even. \\
\textbf{Odd even classification.} 
For a given number $N$ and a string ``even'' or ``odd'', classify if the number satisfies the string condition. \\
\textbf{Decimal operation.} Subtract or divide two numbers. Operands are represented in decimal notation. \\ 
\textbf{Decimal \& word operation.} Subtract or divide two numbers. Operands are represented in decimal or word notation. \\ 
\textbf{Mean.} Given a set of numbers, output the mean.\\ 
\textbf{Median.} Given a set, output the median. \\
\textbf{Mode.} Given a set of numbers, output the mode. 
\subsection{Recognizing regular expressions} \label{regex}
This task paradigm focuses on recognizing regular expressions. The training data consists of positive and negative examples of strings matching a regular expression \cite{bhattamishra-etal-2020-ability}. \\ 
\textbf{Recognize \{0,1,2\}*02*.} 
Recognize if a pattern matches \{0,1,2\}*02*. The maximum length of the patterns is 20.\\
\textbf{Recognize AA*BB*CC*DD*EE*.}
Recognize if a pattern matches AA*BB*CC*DD*EE*. The maximum length of the patterns is 30.

\begin{figure*}[!h]
  \centering
  \begin{subfigure}{\columnwidth}
    \centering
    \includegraphics[width=\linewidth, height=4cm]{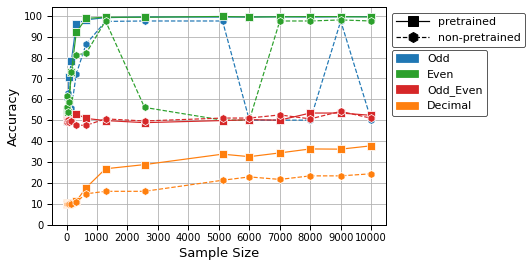}
    \caption{BERT small}
    \label{fig:bert_small_odd_even_oddeven}
  \end{subfigure}%
  \begin{subfigure}{\columnwidth}
    \centering
    \includegraphics[width=\linewidth, height=4cm]{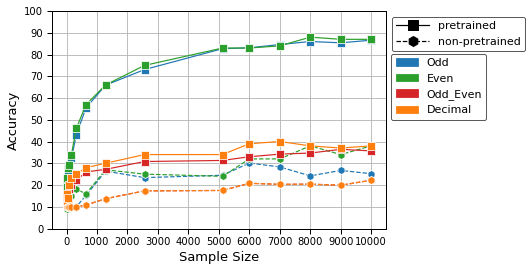}
    \caption{ELMO}
    \label{fig:elmo_dec_wor_decword_odd}
  \end{subfigure}%
  \caption{Performance comparison of pretrained and non-pretrained models of BERT small, and ELMO on four quantitative computation tasks (odd classification, even classification, odd even classification and decimal operation).}
  \label{odd_even_oddeven}
\end{figure*}

\begin{figure*}[!h]
  \centering
  \begin{subfigure}{\columnwidth}
    \centering
    \includegraphics[width=\linewidth, height=4cm]{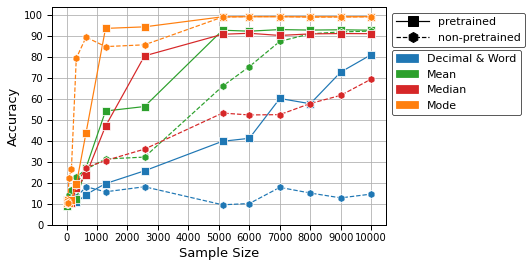}
    \caption{BERT small}
    \label{fig:bertsmall_mea_med_mode}
  \end{subfigure}%
  \hfill
  \begin{subfigure}{\columnwidth}
    \centering
    \includegraphics[width=\linewidth, height=4cm]{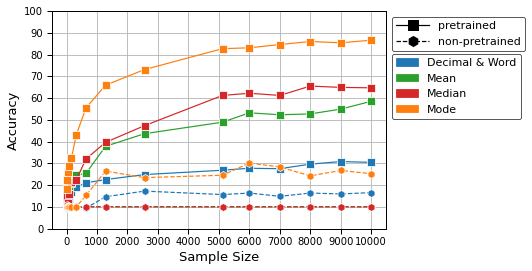}
    \caption{ELMO}
    \label{fig:elmo_mea_med_mod}
  \end{subfigure}%
  \caption{Performance comparison of pretrained and non-pretrained models of BERT small, and ELMO on four quantitative computation tasks (mean, median, mode and decimal \& word operation tasks).}
  \label{mean_med_mod_deciword}
\end{figure*}

\subsection{String reasoning} \label{string}
This task paradigm focuses on reasoning tasks over individual strings or pairs of strings. \\ 
\textbf{Palindrome classification.}
A string is a palindrome if it reads the same forward and backward. The task is to classify whether a given string is a palindrome. The string length ranges from 1 to 15.\\
\textbf{Anagram classification.}
Two strings are anagrams if one is formed by rearranging letters from the other. The task is to classify if a pair of strings are anagrams. The string length ranges from 2 to 15.\\ 
\textbf{Isogram classification.} 
A string is an isogram if it has no repeating characters. The task is to classify whether a given string is an isogram. The string length ranges from 1 to 52.\\
\textbf{Tautonym classification.}
A tautonym is a word which can be broken down into two identical parts, with the same spelling. The task is to classify whether a given string is a tautonym. The string length ranges from 1 to 10. \\
\textbf{Length of a string.}
Output the length of a given string. The string length ranges from 1 to 10.\\
\textbf{Count of unique characters.}
Given a string, count the number of unique characters in it. The string lengths ranges from 10 to 30.\\
\textbf{Parity check.}
Given a binary string, 
output if the counts of ones and zeros are the same. The maximum length of the binary string is 20.\\
\textbf{Vowels classification.}
Given a string, classify if the string contains only vowel characters. The string length ranges from 3 to 10. \\
\textbf{Maximum frequent character.}
Given a string, output the character with the maximum frequency. The string length ranges from 5 to 30. 

\section{Models \& variants} \label{models}
Next, we describe the LMs and their variants used in \nilm. 
We experiment with four language models, based on both Transformer and RNN architectures. 
\textbf{BERT small.}
This is the bert-base-uncased model with 12 transformer encoder layers and the dimension of the representations is 768.  BERT tokenizer is based on the WordPiece model \cite{wu2016google}. \\
\textbf{BERT large.}
This is the bert-large-uncased model which has 24 transformer encoders and representations have 1024 dimensions. \\ 
\textbf{DeBERTa.}
This is a transformer based language model and its tokenizer is built using Byte Pair Encoding \cite{sennrich-etal-2016-neural}. We consider the DeBERTa base model. It has 12 transformer encoder layers and representations have 768 dimensions. \\
\textbf{ELMO.}
This is an LSTM based language model \cite{peters-etal-2018-deep}.
It has 3 layers and the output representations have 1024 dimensions. 

Our experiments are based on pretrained and non-pretrained variants of these architectures. 
For pretrained variants, the weights are initialized with the pretrained weights. The tokenization on the training data is performed using the pre-built vocabulary. For the non-pretrained neural models, the weights are initialized randomly and updated during training. The tokenizer used is the same as in the pretrained variant. 

All the models are trained with varying training data of sizes 10, 20, 40, 80, 160, 320, 640, 1280, 2560, 5120, 6000, 7000, 8000, 9000 and 10000. For training set sizes of less than 1000 samples, we report the average of 10 runs. For training set sizes greater than 1000, all reported numbers are averages of 5 runs. In the next section, we present a comparative analysis of pretrained and non-pretrained models.

\begin{figure*}[!htb]
  \centering
  \begin{subfigure}{\columnwidth}
    \centering
    \includegraphics[width=\linewidth, height=3.8cm]{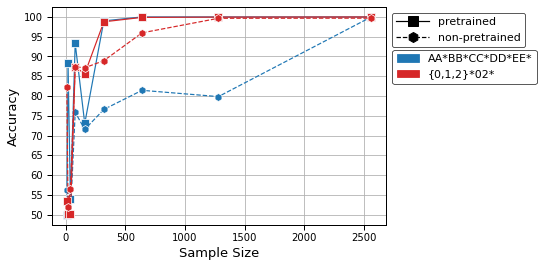}
    \caption{BERT small}
    \label{fig:bertsmall_aab_012_uni}
  \end{subfigure}%
  \hfill
  \begin{subfigure}{\columnwidth}
    \centering
    \includegraphics[width=\linewidth, height=3.8cm]{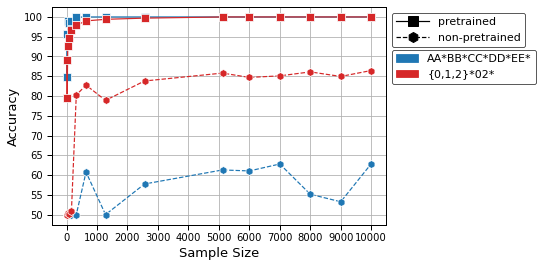}
    \caption{ELMO}
    \label{fig:elmo_aab_012_uni}
  \end{subfigure}%
  \caption{Performance comparison of pretrained and non-pretrained models of BERT small, and ELMO on regular expression tasks (AA*BB*CC*DD*EE* and recognize \{0,1,2\}*02*).}
  \label{aab_012_uni}
\end{figure*}

\section{Comparative Evaluation} \label{results}

Next, we compare the performance of pretrained and non-pretrained models on tasks in \nilm\footnote{Details, including statistical significance results with the paired t-value test, are included in Appendix \ref{tab:pvalmodels}}. 

\noindent \textbf{Quantitative computation}: \label{arthresults}
Figure \ref{odd_even_oddeven} shows results on odd classification, even classification, odd even classification and decimal operation tasks. We find that pretrained LMs outperformed non-pretrained model for all of these tasks. Further, Transformer-based LMs outperformed the RNN-based ELMO models in all the tasks\footnote{We will focus on BERT small as representative of transformer models. Results for BERT large and DeBERTa follow similar trends, and are included in the supplementary material}. We note that for the relatively easy tasks such as odd and even classifications, the pretrained LMs show more stable training. However, for harder tasks such as Decimal operations (where the baseline performance is around 10\%), no models are able to learn the task well even with 10K labeled examples.

Figure \ref{mean_med_mod_deciword} shows results on median, mean, mode and decimal \& word operation tasks. 
The median task requires complex reasoning (sorting numbers and computing the middle element), and shows significantly lower performance than the mean and mode tasks for the non-pretrained models even with the maximum training set size. The pretrained LM models show little eventual difference in performance between these three tasks. On the other hand, for the easiest of these tasks (mode), non-pretrained models actually show higher performance than pretrained LMs in the low data regime.

\noindent \textbf{Recognizing regular expressions:} \label{regexresults}
Figure \ref{aab_012_uni} shows the comparative performance of pretrained LMs on non-pretrained models on the two tasks involving recognizing regular expressions. For both tasks, we note that the pretrained LMs can perfectly learn the tasks with many fewer labeled examples compared to the non-pretrained models. In both cases, the non-pretrained Transformer-based models eventually reach optimal performance as well. However, curiously the ELMO based non-pretrained models struggle with learning both tasks.

\begin{figure*}[!htb]
  \centering
  \begin{subfigure}{\columnwidth}
    \centering
    \includegraphics[width=\linewidth, height=4cm]{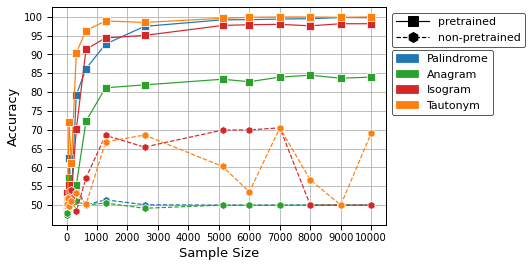}
    \caption{BERT small}
    \label{fig:bertsmall_pal_ana_iso_tau}
  \end{subfigure}%
  \hfill
    \begin{subfigure}{\columnwidth}
    \centering
    \includegraphics[width=\linewidth, height=4cm]{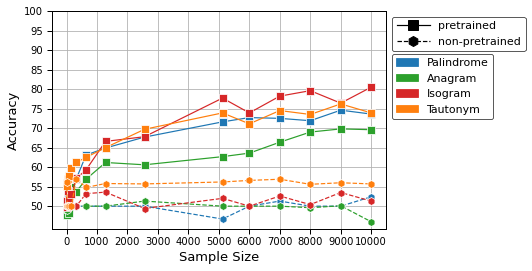}
    \caption{ELMO}
    \label{fig:elmo_pal_ana_iso_tau}
  \end{subfigure}%
  \caption{Performance comparison of pretrained and non-pretrained models of BERT small, and ELMO on four string reasoning tasks (palindrome, anagram, isogram and tautonym classification). }
  \label{pal_ana_iso_tau}
\end{figure*}

\noindent \textbf{String reasoning:} \label{stringresults}
Figures \ref{pal_ana_iso_tau} show the results on Palindrome, Anagram, Isogram and Tautonym classification. These tasks require character comparison within the string or with another string. Again, the pretrained variants consistently outperformed non-pretrained models variants in all of these tasks. In particular, the non-pretrained models completely fail to learn the Anagram and Palindrome tasks even for the largest training set size. Again, Transformer based LMs outperform LSTM based LMs. 

\begin{figure*}[!htb]
  \centering
  \begin{subfigure}{\columnwidth}
    \centering
    \includegraphics[width=\linewidth, height=3.8cm]{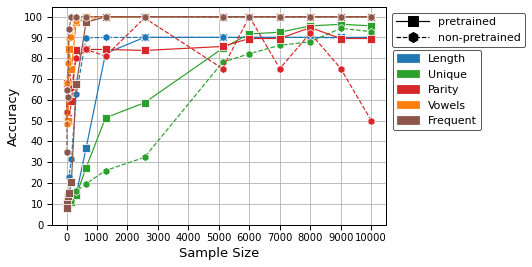}
    \caption{BERT small}
    \label{fig:bertsmall_leng_uniq_par_vow_freq}
  \end{subfigure}%
  \hfill
  \begin{subfigure}{\columnwidth}
    \centering
    \includegraphics[width=\linewidth, height=3.8cm]{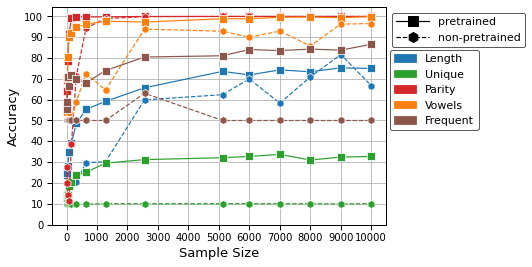}
    \caption{ELMO}
    \label{fig:elmo_len_freq_vow_one}
  \end{subfigure}%
  \caption{Performance comparison of pretrained and non-pretrained models of BERT small, and ELMO on five string reasoning tasks (length of a string, maximum frequent character, vowels classification, parity check and count of unique character).}
  \label{leng_uniq_par_vow_freq}
\end{figure*}

Figure \ref{leng_uniq_par_vow_freq} shows the results on vowels classification, maximum frequent character, length of a string and parity check tasks. These tasks don't require intra-string comparisons. We see that most Transformer-based variants eventually achieve optimal performance. For these simpler tasks, we again observe several instances where the Transformer-based non-pretrained models actually outperform pretrained LMs in the low data regime.  

\section{Effect of model size} \label{param_count}

\begin{figure}[!h]
  \centering
    \includegraphics[width=\linewidth, height=3.8cm]{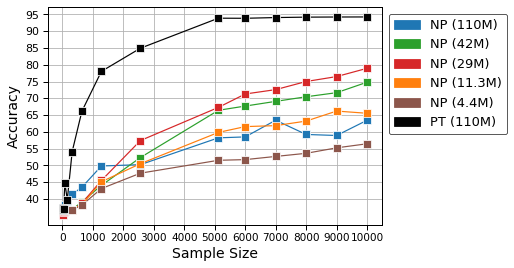}
    \label{fig:bertsmall_model_size}
  \caption{Effect of model size on non-pretrained models. NP denotes a non-pretrained model and PT denotes the pretrained model. Mid-sized non-pretrained models outperform bigger and smaller variants, but still perform significantly lower than pretrained LM models. Results are the average of six representative tasks: palindrome classification, anagram classification, isogram classification, tautonym classification, mean and median. 
  }
  \label{model_size}
\end{figure}

As previously mentioned, a possible explanation for the underperformance of non-pretrained models ise that 
the large number of parameters of the architecture relative to the sizes of the training data might be leading to under-fitting. To test this, we experiment with smaller Transformer-based models with varying numbers of parameters. 

Figure \ref{model_size} illustrates the effect of model sizes of non-pretrained model. The original 110 million parameter model has 12 encoder layers, 12 attention heads, and 768 dimensional representations. The 42 million parameter model has 8 encoder layers, 8 attention heads and 512 dimensional representations. The 29 million parameter model has 4 encoder layers, 8 attention heads and 512 dimensional representations. The 11 million parameter model has 4 encoder layers, 4 attention heads and 256 dimensional representations. The smallest 4 million parameter model has 2 encoder layers, 2 attention heads and 128 dimensional representations. 

As seen in the figure, reducing the model size significantly improves the average performance of the non-pretrained models over 6 representative tasks. However, the smallest models show a performance drop. 
Most significantly, even the best performing intermediate-sized architectures are significantly worse than the pretrained LM models. This strongly suggests that the discrepancy between pretrained and non-pretrained models is not simply due to a mismatch between model and data sizes.

\section{Effects of Pretraining Data} \label{pretrainedablation}
We observe that pretrained LMs consistently performed better than non-pretrained models. This leads to the natural question of what role the text data used for pretraining plays in the process. Next, we investigate this in depth by experimenting with language models pretrained on different types of text. For this, we pretrain models using the BERT-small and DeBERTa architectures and an MLM objective on different text datasets, and evaluate the performance of these models on \nilm tasks. 

\subsection{Variance with text domain}

\begin{table*}[!t]
\scriptsize
    \centering
    \begin{tabular}{c|c|c|c|c|c|c||c||c|c|c||c|c|c||c}
    \hline
         \thead{Sample\\ size} & \thead{SNLI} & \thead{SNLI \\sort} & \thead{SNLI\\ shuffle} & \thead{Amz} & \thead{Amz \\sort} & \thead{Amz\\ shuffle}& \thead{ROC} & \thead{X-ling \\BERT} & \thead{Chinese \\ BERT} & \thead{Code \\ BERT} & \thead{Zipf} & \thead{Unif} &\thead{Syn \\ Voc} & \thead{NP}\\
         \hline
         10 & 37 & 39 & 38 & 36 & 36 & 36 & 36 & 38 & 38 & 37 & 38 & 36 & 36 & 37\\
20 & 37 & 37 & 37 & 36 & 38 & 38 & 38 & 37 & 37 & 38 & 37 & 37 & 37 & 37\\
40 & 37 & 38 & 36 & 37 & 36 & 36 & 36 & 42 & 42 & 37 & 42 & 36 & 37 & 37\\
80 & 38 & 40 & 40 & 37 & 38 & 38 & 38 & 55 & 55 & 47 & 55 & 36 & 36 & 38\\
160 & 38 & 40 & 37 & 37 & 40 & 40 & 40 & 56 & 56 & 37 & 56 & 37 & 37 & 39\\
320 & 40 & 49 & 41 & 38 & 41 & 41 & 41 & 64 & 64 & 61 & 64 & 39 & 37 & 41\\
640 & 44 & 60 & 47 & 43 & 52 & 52 & 52 & 75 & 75 & 69 & 75 & 42 & 39 & 44\\
1280 & 60 & 71 & 63 & 55 & 69 & 69 & 69 & 80 & 80 & 92 & 80 & 52 & 41 & 50\\
2560 & 76 & 84 & 75 & 75 & 79 & 79 & 79 & 81 & 81 & 89 & 81 & 59 & 48 & 50\\
5120 & 82 & 87 & 82 & 83 & 89 & 89 & 89 & 94 & 94 & 97 & 94 & 71 & 58 & 58\\
6000 & 83 & 87 & 83 & 85 & 90 & 90 & 90 & 94 & 94 & 96 & 94 & 73 & 60 & 59\\
7000 & 88 & 89 & 88 & 89 & 91 & 91 & 91 & 94 & 94 & 97 & 94 & 78 & 62 & 64\\
8000 & 89 & 89 & 88 & 90 & 92 & 92 & 92 & 94 & 94 & 97 & 94 & 81 & 63 & 59\\
9000 & 90 & 90 & 89 & 91 & 92 & 92 & 92 & 94 & 94 & 97 & 94 & 84 & 64 & 59\\
10000 & 91 & 88 & 89 & 91 & 92 & 92 & 92 & 94 & 94 & 97 & 94 & 85 & 64 & 64\\

\hline
    \end{tabular}
    \caption{Average accuracy scores of different pretrained BERT representations on six representative non-linguistic tasks: palindrome, anagram, isogram, tautonym, mean, and median.
    The results are rounded to the nearest percentage point.
    All models except Synthetic Vocabulary (Syn Voc). show statistically significant improvements ($p < 0.05$) over the non-pretrained models.}
    \label{tab:nonsyntheticbert}
\end{table*}

We first explore models pretrained on three different domains of text.

\noindent \textbf{SNLI.} We pretrained BERT small from scratch on SNLI data \cite{bowman-etal-2015-large}. It has 1000k sentences (570k pairs of text and hypothesis). \\ 
\noindent \textbf{Amazon reviews.} We selected 500k movies and tv reviews from the larger Amazon reviews dataset \cite{he2016ups} and used for pretraining. Since reviews are in a free-text format, and their collection was not tailored with a NLP task in mind, they might be more representative of the complexity of real-world language use than SNLI.\\
\noindent \textbf{ROC.} ROC is a corpora of 100K children stories, each made up of five sentences \cite{mostafazadeh-etal-2017-lsdsem}. 
The language in ROC is relatively simple in both vocabulary and sentence structure. 

Tables \ref{tab:nonsyntheticbert} and \ref{tab:nonsyntheticdeberta} shows the average accuracy of six non-linguistic tasks (palindrome classification, isogram classification, tautonym classification, odd even classification, decimal operation and median) fine-tuned using different BERT and DeBERTA representations respectively. We note that the models pretrained on all three domains outperformed the non-pretrained model (NP). This suggests that the results of experiments in Section 5 generalize to new text corpora for pretraining, and do not rely on having access to text on specific topics during pretraining. This is a non-trivial result, since it suggests for example, that the higher performance of pretrained models on tasks such as palindrome and anagram classification is not due to the pretrained models having seen information about such concepts during pretraining. This is especially so since the results even generalize to ROC stories, which contain no information on such technical concepts. 

\begin{table*}[!t]
\scriptsize
    \centering
    \begin{tabular}{c|c|c|c|c|c|c||c||c||c|c|c|c}
    \hline
         \thead{Sample\\ size} & \thead{SNLI} & \thead{SNLI \\sort} & \thead{SNLI\\ shuffle} & \thead{Amz} & \thead{Amz \\sort} & \thead{Amz\\ shuffle}& \thead{ROC} & \thead{X-ling \\DeBERTa}  & \thead{Zipf} & \thead{Unif} &\thead{Syn \\ Voc}&  \thead{NP}\\
         \hline
         10 &36 &	36&	37&	36&	35&	36&	37&	36&	37&	36&	36 & 37\\
         20	& 37 &	36	& 36	&36&	35&	35&	37&	39	&36&	37&	37& 37 \\
        40 & 37	& 36&	36&	36&	36&	35&	37&	38&	37&	36&	37 & 37 \\
        80 & 38	&37&	39&	37&	37&	36	&37&	38&	37&	36&	36 & 37 \\
        160 & 37 &	38&	37&	36	&38&	37&	37&	40&	38&	37&	37 & 38 \\
        320 & 39 & 39 & 39 & 37 & 42 & 39 & 41 & 58 & 40 & 39 & 37 & 38 \\
640 & 44 & 44 & 45 & 42 & 52 & 46 & 48 & 71 & 47 & 42 & 39 & 47\\
1280 & 54 & 51 & 54 & 50 & 72 & 58 & 52 & 80 & 61 & 52 & 41 & 60\\
2560 & 70 & 70 & 69 & 65 & 81 & 72 & 65 & 90 & 75 & 59 & 48 & 72\\
5120 & 79 & 78 & 80 & 79 & 87 & 83 & 83 & 93 & 83 & 71 & 58 & 73\\
6000 & 79 & 82 & 80 & 81 & 88 & 84 & 82 & 91 & 84 & 73 & 60 & 74\\
7000 & 84 & 86 & 87 & 85 & 89 & 87 & 84 & 93 & 84 & 78 & 62 & 74\\
8000 & 85 & 87 & 87 & 86 & 89 & 88 & 85 & 93 & 87 & 81 & 63 & 76\\
9000 & 86 & 87 & 88 & 86 & 91 & 90 & 85 & 93 & 88 & 84 & 64 & 77\\
10000 & 87 & 87 & 89 & 86 & 91 & 90 & 85 & 93 & 87 & 85 & 64 & 78\\

\hline
    \end{tabular}
    \caption{Average accuracy scores of different pretrained DeBERTA representations on six representative non-linguistic tasks: palindrome, anagram isogram, tautonym, mean, and median. 
    The results are rounded to the nearest percentage point. 
    All models except Synthetic Vocabulary (Syn Voc). show statistically significant improvements ($p < 0.05$) over the non-pretrained models.}
    \label{tab:nonsyntheticdeberta}
\end{table*}

\subsection{Perturbed text}
Next, we experiment with perturbing the text used for pretraining by changing the order of words in the text. 
We explore the following models:

\noindent \textbf{SNLI sort.} The words in the sentences of SNLI dataset are sorted based on alphabetical order.   \\
\noindent \textbf{SNLI shuffle.} We randomly shuffle words in sentences in the SNLI dataset. \\
\noindent \textbf{Amazon reviews sort.} Similar to SNLI sort, the words in sentences are alphabetically sorted. \\
\noindent \textbf{Amazon reviews shuffle.} We randomly shuffle words in sentences in the Amazon reviews dataset.

\noindent We observe that models pretrained with perturbed text also significantly outperformed non-pretrained models, and perform comparably to the original pretrained LMs. For the SNLI dataset, there is 3\% drop in best performance when pretrained on SNLI sort and 2\% drop in performance when pretrained on SNLI shuffle for BERT (Table 2). In fact, for DeBERTa, SNLI shuffle outperformed the standard SNLI by 2\% (Table 3). Similarly, the Amazon sort and Amazon shuffle versions outperformed or achieved similar performance as the standard Amazon data version. A likely explanation for this is that, even though syntactic word order is disturbed by shuffling, distributional information over sentence contexts is still preserved in the perturbed data.  
We describe experiments with text data having no distributional information in later sections.

\subsection{Non-English and Computer Languages}
A possible rationale for explaining the beneficial effect of pretraining for non-linguistic tasks is that irrespective of whether the tasks require non-linguistic reasoning, their \textit{format} is in language, and hence language models 
should be able to learn these tasks with fewer examples. To test this hypothesis, we also experiment with models pretrained on text from languages different from English, as well as models pretrained on computer code. These include the following models: \\ 
\textbf{Multilingual BERT.}
Multilingual BERT 
is pretrained on text from 102 different languages. About 21\% of the pretraining text is English.\\
\textbf{Chinese BERT.}
Chinese BERT 
is a BERT model pretrained on Chinese text. \\
\textbf{Code BERT.}
CodeBERT \cite{feng-etal-2020-codebert} is pretrained on code from six programming languages.

In Table~\ref{tab:nonsyntheticbert}, we note that all three non-English pretrained LMs significantly outperformed non-pretrained models, with the best performance being comparable or marginally lower than English versions. In fact, Code-BERT surprisingly surpasses ROC by 5\%. These findings strongly indicate that the advantages from pretraining have little to do with the format of the tasks, since they persist for scenarios with little shared linguistic structure. 

\subsection{Synthetic languages}
Finally, to investigate what happens if we weaken the distributional properties that hold even in the perturbed text versions from Section 6.2, we experiment with pretraining models on synthetic text sampled from simple probability distributions: 

\noindent \textbf{Zipf distribution.} We select 30k words (types) from the Amazon reviews dataset. Words are picked with a unigram probability that follows Zipf's word frequency law, which all natural languages empirically follow~\cite{piantadosi2014zipf}. For the Zipf distribution, we chose $\alpha$=1 and $\beta$=2.7, to match the parameters of most natural languages. The text does not follow any word order.\\
\noindent \textbf{Uniform distribution.} In this dataset, words are sampled from the same vocabulary as in `Zipf distribution', but with a uniform unigram probability. The text does not follow any word order.\\
\noindent \textbf{Synthetic Vocabulary.} Words are selected with uniform distribution from a vocabulary to form sentences. However, instead of a vocabulary of English words, the words in the vocabulary are also synthetically generated (3 letter combinations of lower-case alphabets).  In this text, the words do not possess morphology in addition to no syntax.

In Tables ~\ref{tab:nonsyntheticbert} and ~\ref{tab:nonsyntheticdeberta}, we note that surprisingly, even models pretrained on Zipfian and uniform distribution text continue to outperform the non-pretrained models. In fact, the Zipf version's best accuracy is 3\% higher than the standard Amazon data version and 2\% compared to perturbed Amazon shuffled data version in case of BERT. Zipf outperforms standard amazon data by 1\% and lags behind amazon shuffle by 3\% for DeBERTA. The Uniform distribution version lags behind Zipf by 9\% and 2\% for BERT and DeBERTa respectively. We note that the Zipf and Uniform versions still use the prebuilt vocabulary from the Amazon data, and hence this text maintains morphological structure. 
However, the gains finally disappear for the Synthetic vocabulary model, which cannot leverage morphological structure in the text, and its performance is similar to the non-pretrained models.

\section{Conclusion}
We explore the non-linguistic inductive biases of pretrained LMs. 
While the general trend (that pretraining helps) is unsurprising, our analysis with models pretrained on different text corpora 
shows that this is not due to the model seeing related topics during pretraining. We find that these gains persist even in absence of any shared linguistic structure (in cross-lingual settings). 
Our observation that this behavior is seen even when pretraining on synthetically generated languages is intriguing and can be explored further by future work. 

\section*{Acknowledgements}
This work was supported in part by NSF grant DRL2112635. We are also thankful to the anonymous reviewers for their thoughtful suggestions.

\section*{Ethics and Broader Impact}
Our synthetic datasets contain no linguistic or social information, and hence cannot introduce any type of social, gender and cultural biases in our analyses. The datasets used in the section \ref{pretrainedablation} are publicly available, and should contribute towards the goal of reproducible research. In terms of broader impact, our results suggest that LMs accrue helpful inductive biases for non-linguistic reasoning during pretraining. This suggests that LMs can potentially be explored for a broader range of downstream applications rather than language-related tasks, which is the current predominant focus of these models. In the long run, making such foundational models available for learning a broad range of tasks from limited data can make predictive AI technologies more accessible than in the current day.

\section*{Limitations}
In terms of findings, we find strong evidence of pretraining on text providing advantageous inductive biases for non-linguistic tasks. Our analysis in Section 6 suggests that this is not simply a regularization effect. However, it does not definitively rule out this possibility since direct comparisons between pretrained and non-pretrained networks (even of different sizes) are difficult.
Also, the scope of our analysis here is limited to small to mid-sized language models (with tens of millions of parameters), rather than massive language models such as GPT3 (with tens of billions of parameters). 
Finally, we note that all tasks chosen for this analysis are formulated as classification, where the number of classes is not high. Hence, learning some of the tasks might easier than possible more general formulations. e.g., quantitative computation. 
\bibliography{anthology,custom}
\bibliographystyle{acl_natbib}

\clearpage
\appendix
\counterwithin{figure}{section}
\section{Appendix}
\begin{table}[!htp]
    \centering
    \begin{tabular}{|c|c|}
    \hline
       \thead{Baseline}  & \thead{p-value} \\
       \hline
       SNLI  & $5.45 \times 10^{-5}$\\
       SNLI sort & $3.33 \times 10^{-4}$\\
       SNLI shuffle & $5.5 \times 10^{-4}$ \\
       Amazon & $7.48 \times 10^{-5}$ \\
       Amazon sort & $7.2 \times 10^{-5}$ \\
       Amazon shuffle & $4.5 \times 10^{-5}$ \\
       Multilingual BERT & $9.07 \times 10^{-4}$ \\
       Chinese BERT & $8.9 \times 10^{-5}$ \\
       Code BERT & $8.1 \times 10^{-5}$ \\
       ROC & $2.64 \times 10^{-5}$\\
       Zipf distribution & $7.45 \times 10^{-5}$ \\
       Uniform distribution & $4.61 \times 10^{-4}$ \\
       Synthetic vocabulary & $1.2 \times 10^{-1}$ \\
       \hline
    \end{tabular}
    \caption{Statistical significance values (paired t-test) between non-pretrained model and other baseline BERT models trained on different datasets.}
    \label{tab:pvaldatabert}
\end{table}

\begin{table}[!htp]
    \centering
    \begin{tabular}{|c|c|}
    \hline
       \thead{Baseline}  & \thead{p-value} \\
       \hline
       SNLI  & $2.45 \times 10^{-5}$\\
       SNLI sort & $1.33 \times 10^{-4}$\\
       SNLI shuffle & $4.3 \times 10^{-5}$ \\
       Amazon & $6.32 \times 10^{-4}$ \\
       Amazon sort & $8.7 \times 10^{-5}$ \\
       Amazon shuffle & $7.3 \times 10^{-5}$ \\
       Multilingual BERT & $9.07 \times 10^{-5}$ \\
       ROC & $2.14 \times 10^{-3}$\\
       Zipf distribution & $3.1 \times 10^{-3}$ \\
       Uniform distribution & $4.61 \times 10^{-4}$ \\
       Synthetic vocabulary & $1.3 \times 10^{-1}$ \\
       \hline
    \end{tabular}
    \caption{Statistical significance values (paired t-test) between non-pretrained model and other baseline DeBERTA models trained on different datasets.}
    \label{tab:pvaldatadeberta}
\end{table}

\begin{figure*}[!htbp]
  \centering
  \begin{subfigure}{\columnwidth}
    \centering
    \includegraphics[width=\linewidth, height=3.8cm]{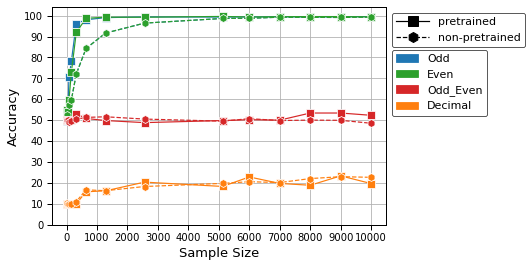}
    \caption{DeBERTa}
    \label{fig:deberta_dec_wor_decword_odd}
  \end{subfigure}%
  \hfill
  \begin{subfigure}{\columnwidth}
    \centering
    \includegraphics[width=\linewidth, height=3.8cm]{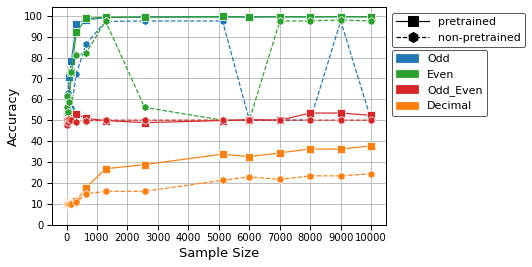}
    \caption{BERT large}
    \label{fig:bertlargepal_dec_wor_decwor_odd}
  \end{subfigure}%
  \caption{Performance comparison of pretrained and non-pretrained models of  DeBERTa and BERT large on four quantitative computation tasks (odd classification, even classification, odd even classification and decimal operation).}
  \label{dec_wor_decword_odd2}
\end{figure*}
\begin{figure*}[!htbp]
  \centering
  \begin{subfigure}{\columnwidth}
    \centering
    \includegraphics[width=\linewidth, height=4cm]{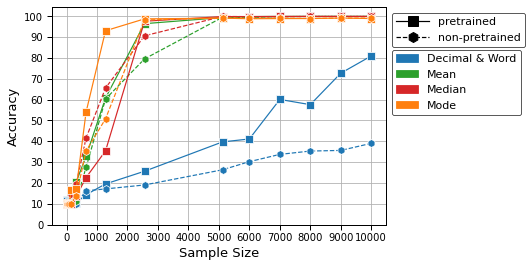}
    \caption{DeBERTa}
    \label{fig:deberta_mea_med_mod}
  \end{subfigure}%
  \hfill
  \begin{subfigure}{\columnwidth}
    \centering
    \includegraphics[width=\linewidth, height=4cm]{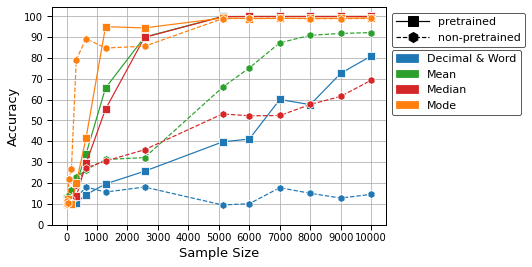}
    \caption{BERT large}
    \label{fig:bertlargepal_mea_med_mod}
  \end{subfigure}%
  \caption{Performance comparison of pretrained and non-pretrained models of DeBERTa and BERT large on four quantitative tasks (mean, median, mode, decimal \& word operation).}
  \label{mea_med_mod2}
  \end{figure*}
  
  \begin{figure*}[!htbp]
  \centering
  \begin{subfigure}{\columnwidth}
    \centering
    \includegraphics[width=\linewidth, height=3.8cm]{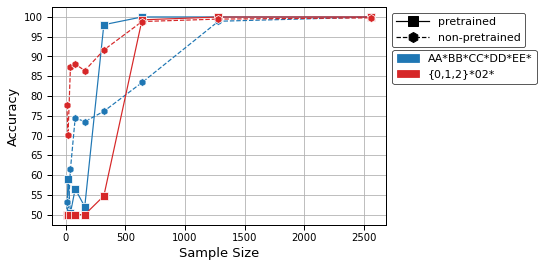}
    \caption{BERT small}
    \label{fig:bertsmall_aab_012_uni}
  \end{subfigure}%
  \hfill
  \begin{subfigure}{\columnwidth}
    \centering
    \includegraphics[width=\linewidth, height=3.8cm]{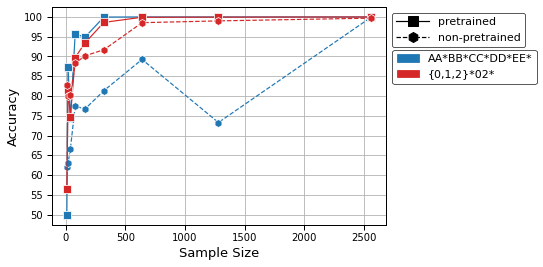}
    \caption{ELMO}
    \label{fig:elmo_aab_012_uni}
  \end{subfigure}%
  \caption{Performance comparison of pretrained and non-pretrained models of  DeBERTa, and  BERT large on regular expression tasks (AA*BB*CC*DD*EE* and recognize \{0,1,2\}*02*).}
  \label{aab_012_uni2}
\end{figure*}
\begin{figure*}[!htbp]
  \begin{subfigure}{\columnwidth}
    \centering
    \includegraphics[width=\linewidth, height=4cm]{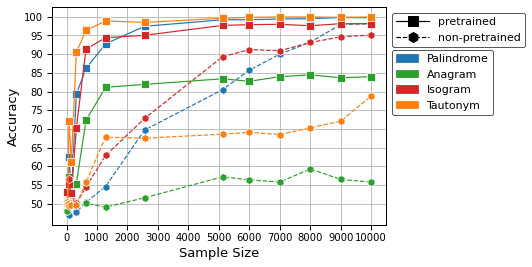}
    \caption{DeBERTa}
    \label{fig:deberta_pal_ana_iso_tau}
  \end{subfigure}%
  \hfill
  \begin{subfigure}{\columnwidth}
    \centering
    \includegraphics[width=\linewidth, height=4cm]{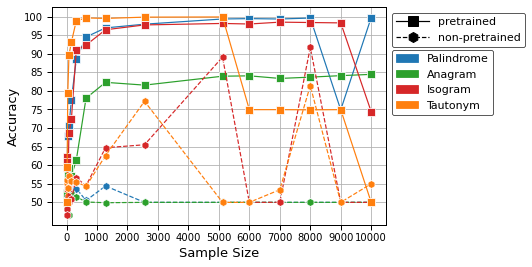}
    \caption{BERT large}
    \label{fig:bertlargepal_ana_iso_tau}
  \end{subfigure}%
  \caption{Performance comparison of pretrained and non-pretrained models of DeBERTa and BERT large on four string reasoning (palindrome, anagram, isogram and tautonym classification). }
  \label{pal_ana_iso_tau2}
\end{figure*}
\begin{figure*}[!htbp]
  \begin{subfigure}{\columnwidth}
    \centering
    \includegraphics[width=\linewidth, height=3.8cm]{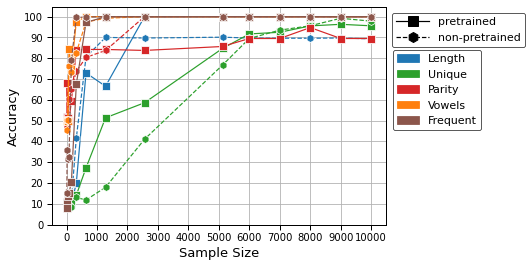}
    \caption{DeBERTa}
    \label{fig:deberta_len_freq_vow_one}
  \end{subfigure}%
  \hfill
  \begin{subfigure}{\columnwidth}
    \centering
    \includegraphics[width=\linewidth, height=3.8cm]{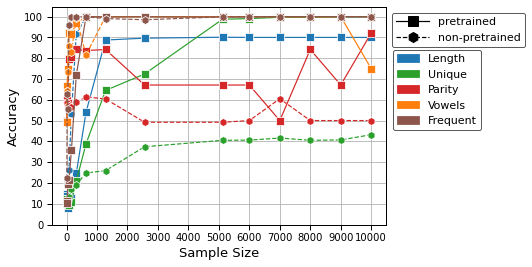}
    \caption{BERT large}
    \label{fig:bertlarge_len_freq_vow_one}
  \end{subfigure}%
  \caption{Performance comparison of pretrained and non-pretrained models of DeBERTa and BERT large on five string reasoning tasks (length of a string, maximum frequent character, vowels classification, parity check and count of unique character).}
  \label{len_freq_vow_one2}
\end{figure*}
\begin{table*}[!htbp]
    \centering
    \begin{tabular}{|c|c|c|c|c|}
    \hline
    \thead{Task} & \thead{BERT small} & \thead{DeBERTa} & \thead{BERT large} & \thead{ELMO} \\
    \hline
    Odd classification & $10.4 \times 10^{-2}$ & $8.8 \times 10^{-1}$& $2.9 \times 10^{-3}$& $7.35 \times 10^{-7}$\\
    
    Even classification & $8.1 \times 10^{-2}$ & $8.7 \times 10^{-2}$& $5.25 \times 10^{-3}$& $7.35 \times 10^{-7}$\\
    
    Odd even classification & $2.2 \times 10^{-1}$& $6.96 \times 10^{-7}$& $6.46 \times 10^{-4}$ & $7.35 \times 10^{-7}$\\
    
    Decimal operation & $4.1 \times 10^{-4}$ & $7.07 \times 10^{-1}$ & $1.35 \times 10^{-5}$ & $3.49 \times 10^{-7}$\\

    Decimal \& word operation & $6.85 \times 10^{-8}$& $6.43 \times 10^{-7}$ & $4.34 \times 10^{-8}$ & $5.39 \times 10^{-7}$\\

    Mean &$9.5 \times 10^{-2}$ & $7.56 \times 10^{-1}$ &$7.8 \times 10^{-6}$ & $2.2 \times 10^{-7}$\\

    Median & $9.28 \times 10^{-6}$& $8.04 \times 10^{-1}$ &$5.68 \times 10^{-7}$ & $1.99 \times 10^{-7}$\\

    Mode & $9.2 \times 10^{-2}$& $2.27 \times 10^{-1}$ & $9.2 \times 10^{-1}$& $3.35 \times 10^{-7}$\\

    Recognize \{0,1,2\}*02* & $1.31 \times 10^{-1}$&$8.4 \times 10^{-1}$ & $4.34 \times 10^{-1}$& $5.48 \times 10^{-5}$\\

    Recognize AA*BB*CC*DD*EE* &$4.06 \times 10^{-1}$ &$6.97 \times 10^{-1}$ & $4.02 \times 10^{-1}$& $2.39 \times 10^{-6}$\\

    Palindrome classification & $4.34 \times 10^{-7}$ & $2.1 \times 10^{-3}$ &$1.85 \times 10^{-7}$ & $1.97 \times 10^{-6}$\\

    Anagram classification & $5.1 \times 10^{-6}$ & $1.44 \times 10^{-6}$ &  $3.45 \times 10^{-7}$& $7.46 \times 10^{-6}$\\

    Isogram classification & $1.28 \times 10^{-7}$ & $4.77 \times 10^{-3}$& $3.47 \times 10^{-4}$& $2.18 \times 10^{-6}$\\

    Tautonym classification & $1.92 \times 10^{-7}$ & $1.29 \times 10^{-5}$ & $1.69 \times 10^{-8}$ & $4.39 \times 10^{-6}$ \\

    Length of a string & $2.7 \times 10^{-1}$& $1.27 \times 10^{-4}$& $3.39 \times 10^{-4}$& $7.07 \times 10^{-4}$\\

    Count of unique characters &$1.79 \times 10^{-4}$ & $2.7 \times 10^{-2}$&$1.23 \times 10^{-7}$ & $3.18 \times 10^{-6}$\\

    Parity check & $2.68 \times 10^{-4}$& $4.66 \times 10^{-4}$ &$4.34 \times 10^{-7}$ & $6.05 \times 10^{-6}$\\

    Vowels classification & $4.26 \times 10^{-1}$ & $9.5 \times 10^{-1}$ & $7.22 \times 10^{-1}$& $5.11 \times 10^{-2}$\\

    Maximum frequent character & $5.02 \times 10^{-1}$ & $5.65 \times 10^{-1}$ & $6.07 \times 10^{-1}$ & $6.47 \times 10^{-1}$\\
    \hline
    \end{tabular}
    \caption{Statistical significance values (paired t-test) between pretrained and non-pretrained model on all the tasks.}
    \label{tab:pvalmodels}
\end{table*}
\subsection{Implementation details}
For transformer LMs, we add a fully connected classification layer on the top of final encoder layer. The pooled representations from the final encoder layer are then passed onto fully connected layer. We train these models in an end-to-end manner. For the RNN LMs, we first pretrain LM onto the task. The final word representations are the weighted sum of three layers. Max-pooling operation is applied on the time step dimension for these weighted representations. A final classification layer is trained with the pooled representations.
\subsection{Computational requirements}
All the models are run using PyTorch framework on 4 geforce gtx 1080 gpus. Each of the fine-tuning experiments takes about 5 gpu hours and pre-training takes about 10 gpu hours.
\subsection{Statistical significance}
We perform a paired t-test between pretrained and non-pretrained models of the LMs on all the tasks. The statistical significance values are shown in the table \ref{tab:pvalmodels}. We also calculated the paired t-value between non-pretrained model and BERT and DeBERTA pretrained on different datasets. The paired t-values are shown in the table \ref{tab:pvaldatabert} and \ref{tab:pvaldatadeberta}. 

\end{document}